\documentclass[runningheads]{llncs}

\usepackage[T1]{fontenc}
\usepackage{verbatim}
\usepackage{graphicx}
\usepackage{xspace}
\usepackage{caption}
\usepackage{xcolor}
\usepackage{booktabs} %
\usepackage{graphicx} %
\usepackage{multirow} %
\usepackage{amsmath}
\usepackage{amssymb}
\usepackage{wrapfig} 
\usepackage{booktabs}
\usepackage{hyperref}
\usepackage{cleveref} %

\newcommand{\name}{TopoOR\xspace}
\newcommand{\modelname}{HAT\xspace}

\begin{document}
\title{TopoOR: A Unified Topological Scene Representation for the Operating Room}
\titlerunning{TopoOR}
\author{Tony Danjun Wang\inst{1,2}\thanks{Corresponding author: \email{tony.wang@tum.de}} \and
Ka Young Kim\inst{4}\and
Tolga Birdal\inst{3} \and \\
Nassir Navab \inst{1,2} \and
Lennart Bastian \inst{1,2,3}}
\authorrunning{T.D. Wang et al.}
\institute{Technical University of Munich, Munich, Germany\and
Munich Center for Machine Learning, Germany \and
Imperial College London, United Kingdom \and
Kyung Hee University, South Korea
}

\maketitle
\begin{abstract}

Surgical Scene Graphs abstract the complexity of surgical operating rooms (OR) into a structure of entities and their relations, but existing paradigms suffer from strictly dyadic structural limitations. 
Frameworks that predominantly rely on pairwise message passing or tokenized sequences flatten the manifold geometry inherent to relational structures and lose structure in the process. 
We introduce \name, a new paradigm that models multimodal operating rooms as a higher-order structure, innately preserving pairwise and group relationships. 
By lifting interactions between entities into higher-order topological cells, \name natively models complex dynamics and multimodality present in the OR. 
This topological representation subsumes traditional scene graphs, thereby offering strictly greater expressivity. %
We also propose a higher-order attention mechanism that explicitly preserves manifold structure and modality-specific features throughout hierarchical relational attention. 
In this way, we circumvent combining 3D geometry, audio, and robot kinematics into a single joint latent representation, preserving the precise multimodal structure required for safety-critical reasoning, unlike existing methods.
Extensive experiments demonstrate that our approach outperforms traditional graph and LLM-based baselines across sterility breach detection, robot phase prediction, and next-action anticipation.

\end{abstract}

\section{Introduction}
\vspace{-1.0mm}
\label{sec:introduction}

\begin{figure}[ht]
    \centering
    \includegraphics[width=\textwidth]{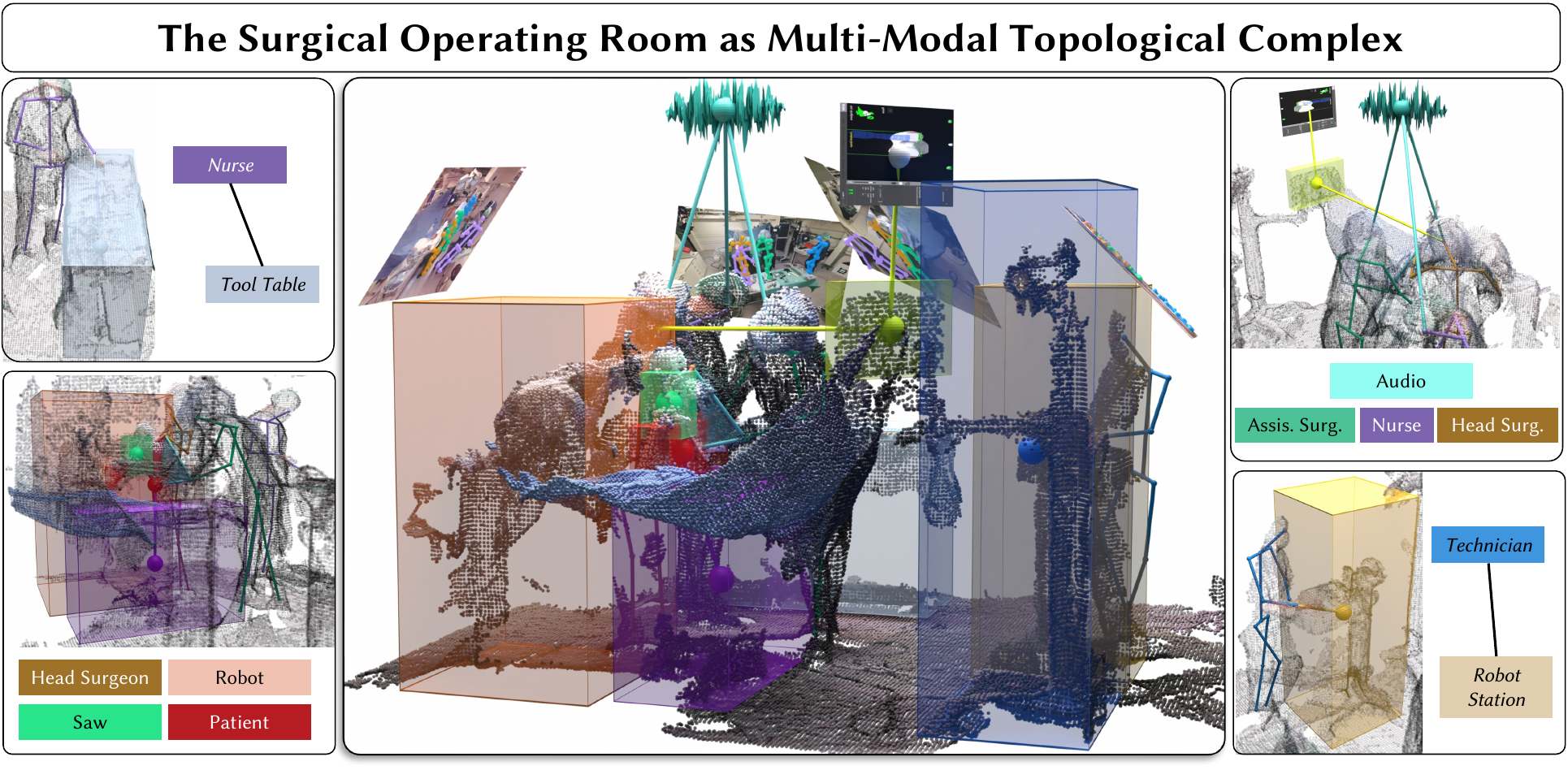}
    \caption{We model surgical ORs as higher-order structures. 
    Our framework explicitly instantiates \textit{cells} anchored in 3D for physical entities—such as the nurse and surgical robot—as well as data sources from diverse modalities, including audio.
    Combining these elements into a higher-order structure enables modeling of the unified interaction among the head surgeon, robot, saw, and patient (bottom left), while preserving the complex, multi-actor dynamics of surgical procedures.}
    \label{fig:teaser}
    \vspace{-4.5mm}
\end{figure}

The overarching goal of Surgical Data Science (SDS) is to improve patient outcomes and procedural efficiency by developing computational models of the surgical operating room (OR) \cite{bigdelou2011or,lalys2014surgical,maier2017surgical}.
Early work applied machine learning models \cite{padoy2019machine,kennedy2020computer} for tasks such as endoscopic phase recognition \cite{twinanda2016endonet,garrow2021machine}, skill assessment \cite{vedula2016analysis}, and tool tracking \cite{bouget2017vision}.
Moving beyond these isolated tasks, action triplets model joint interactions between an instrument and target through an action (e.g., $\langle$ grasper, retract, gallbladder $\rangle$)  \cite{katic2014knowledge,sharma2023rendezvous}.
Surgical scene graphs (SSGs) were later conceived to formally structure the complexity of the entire operating theater by encoding relationships among humans, tools, and equipment into a structured, holistic representation \cite{johnson2015image,islam2020learning,ozsoy2021multimodal}.
However, a significant gap remains between their theoretical ambition and practical implementation.

We posit, for the first time, that surgical procedures are irreducibly \textit{polyadic}.
For instance, during a robotic-assisted bone resection \cite{ozsoy2025mm}, the head surgeon dynamically guides a robotic arm and surgical saw to resect the patient's anatomy while continuously adjusting based on visual feedback from a navigation monitor (see \cref{fig:teaser}).
Standard graphs artificially fragment this unified loop into isolated \textit{dyadic} links (e.g., Surgeon--Robot, Surgeon--Monitor, Saw--Patient), stripping away the joint spatial and kinematic constraints that collectively describe the interaction \cite{wang2025beyond}.
The resulting representation is strictly less expressive than one that models it natively \cite{hajij2022topological}, with potential downstream impacts on workflow understanding and ultimately patient safety.

Another limitation is semantic.
The multimodal data that characterize the OR---including articulated 3D human motion in $\mathrm{SE}(3)$ \cite{wang2026compose}, robot joint kinematics \cite{ahmidi2017dataset}, audio spectrograms \cite{elizalde2023clap}, and RGB appearance features---each reside on geometric manifolds with distinct characteristics.
The relationships among these entities inherit this geometry: the space of valid surgical scene configurations forms a high-dimensional, coupled data manifold that \textit{cannot} be faithfully represented in a single latent space while preserving its manifold structure.
VLM-based approaches attempt to sidestep explicit graph construction by mapping heterogeneous data to a shared latent representation \cite{ozsoy2025mm}, but tokenizing forces this inherently non-Euclidean data through a semantic bottleneck that discards its metric and topological structure.
As a result, these models flatten the manifold geometry of the surgical scene, stripping away structure essential to downstream clinical tasks and surgical safety \cite{hashimoto2018artificial}.

To overcome these limitations, we propose a generalized framework for surgical reasoning rooted in algebraic topology, designed to adequately model the complexity of the operating room.
Rather than relying on pairwise interactions between entities, we model the scene as a \textit{combinatorial complex} (CC) \cite{hajij2022topological,hajij2023combinatorial}, preserving the structural and semantic meaning of higher-order relationships.
To reason over these structures, we employ higher-order attention networks (\modelname) which distribute and aggregate messages directly across the higher-order incidence structure of the complex, establishing a learned hierarchical representation of the surgical scene.
We evaluate our framework on the multimodal MM-OR dataset, jointly optimizing next-action anticipation and robot-phase prediction while enabling zero-shot, rule-based sterility breach detection.
We further show that our topological representation subsumes traditional approaches: when supervised with flattened scene graph labels, it predicts the conventional tokenized format more effectively than existing baselines.
Our primary contributions can be summarized as follows:

\begin{itemize}
    \item \textbf{\textsc{\name}}: We introduce a unified topological framework for surgical scenes, modeling the operating room 
    as a higher-order structure.
    Our higher-order attention mechanism preserves multimodal geometry and captures dynamics between entities without losing structure and semantic meaning.
    \item \textbf{Empirical Performance}: \textsc{\name} outperforms 
    existing graph methods on the MM-OR dataset through a higher-order attention (\modelname) mechanism that preserves structure essential for downstream clinical tasks.
    \item \textbf{Expressiveness}: We demonstrate that our higher-order representation subsumes traditional scene graphs by optimizing directly for downstream tasks while also decoding flattened tokenized relationships.
\end{itemize}

\section{Methodology}
\vspace{-1.5mm}
\label{sec:methodology}

\begin{figure}[ht]
    \centering
    \includegraphics[width=\textwidth]{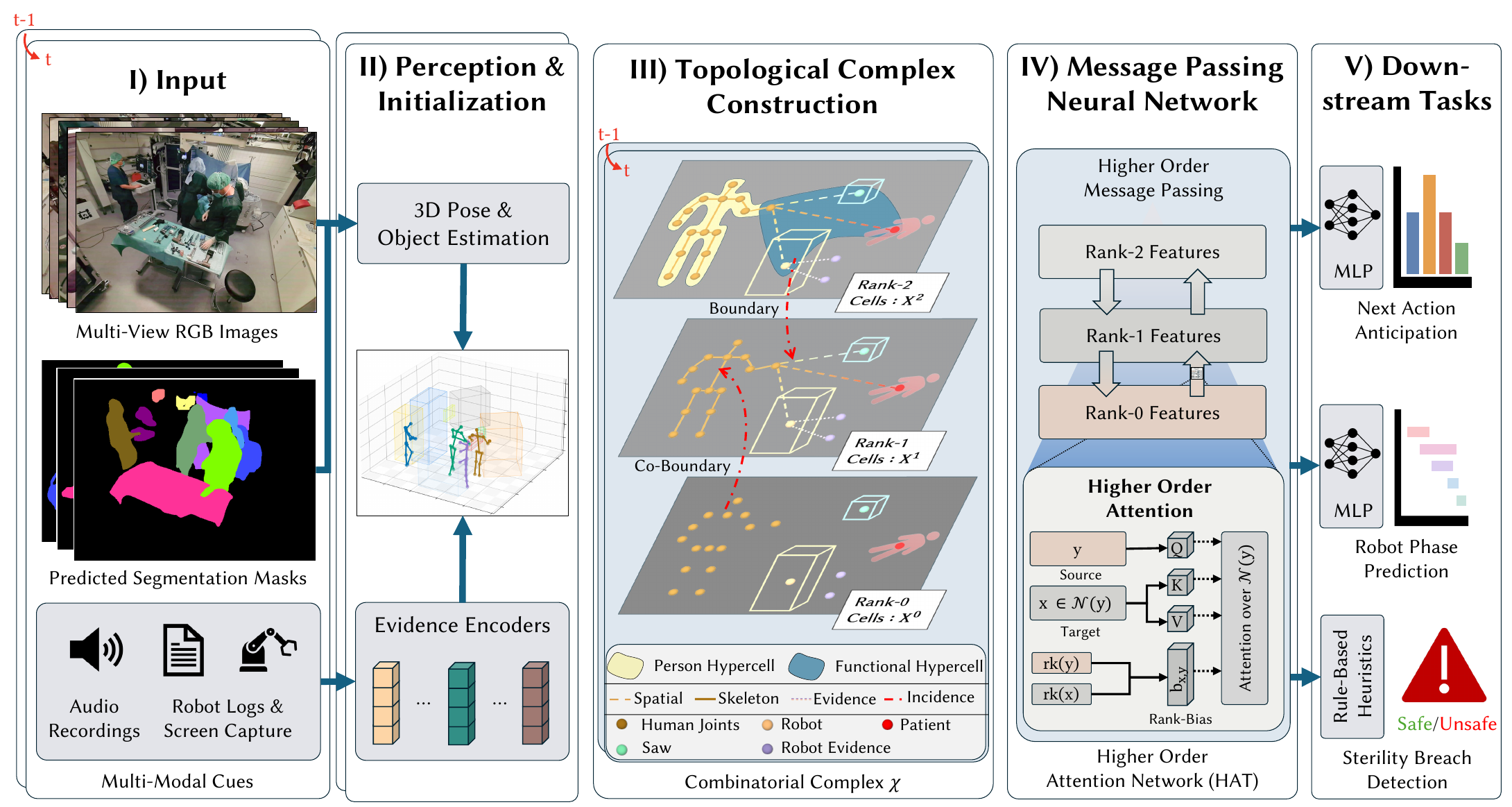}
    \caption{Overview of \name. 
    Given multi-modal sensory inputs over a temporal window (I), 3D entities and evidence features are initialized (II) and abstracted into a CC $\mathcal{X}$ (III). 
    Higher-order attention is computed across incidence neighborhoods $x \in \mathcal{N}(y)$, using a learnable rank-bias to preserve structural heterogeneity (IV). 
    Pooled representations are routed to downstream tasks enabling, e.g., simultaneous next-action anticipation and robot-phase prediction (V).}
    \label{fig:pipeline}
    \vspace{-4.5mm}
\end{figure}

To overcome the limitations of graph-based surgical domain models, we model the collective relationships between individuals and entities in the OR as a higher-order topological structure that encapsulates heterogeneous multi-actor interactions.
To this end, we formulate the preliminaries of our approach (\cref{subsec:preliminaries}), define our Higher-Order Attention Network (\modelname) (\cref{subsec:hot}), detail the construction of the multimodal combinatorial complex (\cref{subsec:graph_construction}), and finally outline the implementation details and downstream multi-task inference (\cref{subsec:implementation}).
A visual overview is provided in \cref{fig:pipeline}.

\vspace{-2.5mm}
\subsection{Preliminaries}
\label{subsec:preliminaries}

To model the heterogeneous nature of the OR in a unified manner, we abstract the surgical scene using the topological deep learning (TDL) framework \cite{hajij2022topological,papamarkou2024position}.

\begin{definition}[Combinatorial Complex~\cite{hajij2023combinatorial}]
A combinatorial complex (CC) consists of a finite set of cells $X$ and a rank function $\text{rk}: X \to \mathbb{Z}_{\geq 0}$ which partitions the complex into $k$-dimensional sets $X^k$. 
The CC is endowed with a partial ordering $\preceq$ denoting the face relation, where $x \preceq y$ indicates that cell $x \in X^p$ is on the boundary of cell $y \in X^q$ with $p < q$.

\end{definition}

CCs generalize graphs and other higher order structures by combining features of cell complexes (hierarchy among relationships) and hypergraphs (arbitrary set-type relations) \cite{hajij2023combinatorial}. 
Consider, for instance, that a graph is a CC with rank-0 nodes $X^0$ connected by rank-1 edges $X^1$. 
To realize this, let us define how cells of different dimensions interact through incidence relations:

\begin{definition}[Incidence Neighborhoods]
For any cell $y \in X^k$, we define its \textit{boundary neighborhood} 
$\mathcal{B}(y) = \{x \in X^p \mid x \preceq y,\; p < k\}$ and its 
\textit{co-boundary neighborhood} $\mathcal{C}(y) = \{z \in X^q \mid y \preceq z,\; q > k\}$. 
The full incidence neighborhood is their union: 
$\mathcal{N}(y) = \mathcal{B}(y) \cup \mathcal{C}(y)$.
\end{definition}

\subsection{Higher-Order Attention Network (\modelname)}
\label{subsec:hot}

Next, we introduce Higher-Order Attention Networks (\modelname), which generalizes GAT \cite{velivckovic2017graph} from graph neighborhoods to the incidence structure of combinatorial complexes \cite{chen2020hypergraph,hajij2022higher,hajij2022topological}, as an inductive bias over higher-order surgical scenes.

\begin{definition}[Higher-Order Attention Layer]
Let $\mathcal{X}$ be a CC with cell features 
$\{h_y^{(l)}\}_{y \in X}$ at layer $l$. For each cell 
$y \in X^k$, \modelname computes:
\begin{equation}
    h_y^{(l+1)} = h_y^{(l)} + 
    \sum_{x \in \mathcal{N}(y) \cup \{y\}} 
    \alpha_{yx}^{(l)}\, W_V^{(l)}\, h_x^{(l)},
\end{equation}

where the attention coefficients are given by
\begin{equation}
    \alpha_{yx}^{(l)} = 
    \underset{x' \in \mathcal{N}(y) \cup \{y\}}{\mathrm{Softmax}}
    \left(
    \frac{
        \langle\, W_Q^{(l)} h_y^{(l)},\; W_K^{(l)} h_x^{(l)} \rangle
    }{\sqrt{d_k}} 
    + b_{\mathrm{rk}(y),\, \mathrm{rk}(x)}
    \right),
\end{equation}
with rank-pair bias
\begin{equation}
    b_{\mathrm{rk}(y),\, \mathrm{rk}(x)} = 
    \phi\!\left(\, 
        \mathbf{e}_{\mathrm{rk}(y)} \odot \mathbf{e}_{\mathrm{rk}(x)} 
    \,\right),
\end{equation}
where $\mathbf{e}_r \in \mathbb{R}^{d_r}$ are learnable rank 
embeddings and $\phi: \mathbb{R}^{d_r} \to \mathbb{R}^{H}$ is a 
linear projection producing per-head biases.
\end{definition}

Information in \modelname flows primarily along the incidence 
structure of $\mathcal{X}$: boundary cells 
($\mathrm{rk}(x) < \mathrm{rk}(y)$) propagate entity-level 
features upward, while co-boundary cells 
($\mathrm{rk}(x) > \mathrm{rk}(y)$) distribute aggregated 
group context downward.
The rank-pair bias $b_{\mathrm{rk}(y),\, \mathrm{rk}(x)}$ modulates information flow based on the topological relationship between source and target cells.
For nodes of distinct rank in $\mathcal{X}$, we can use features from various modalities, detectors, or sensors. 
By exchanging information between these heterogeneous spaces based on the rank function $\text{rk}(x)$, \modelname retains the structural origin and information of each feature. 
This allows the model to differentiate between, e.g., human kinematics modelled on $X^0$ and aggregated multi-actor behavior in $X^2$.

\subsection{Multimodal Graph Construction}
\label{subsec:graph_construction}

To construct our higher-order surgical scene representation, we build the CC $\mathcal{X}$ by using spatial thresholding and imposing clinical priors.

\noindent\textbf{Rank-0 Cells ($X^0$) for Physical Entities and Evidence.} 
These atomic nodes represent human anatomical joints (acquired via 3D pose estimation with semantic embeddings) and 3D-localized object entities (initialized via 3D spatial boundaries). 
We also establish \textit{auxiliary evidence nodes} to integrate additional modalities, including processing robot logs with a language model \cite{reimers2019sentence}, monitoring screens with a small CNN, and spatial audio with an audio encoder \cite{elizalde2023clap}.

\noindent\textbf{Rank-1 Cells ($X^1$) for Interactions.}
We construct intra-entity skeleton edges using predefined human kinematic trees, and inter-entity spatial edges formed dynamically when physical proximity falls below a threshold. 
We further enforce predefined domain-specific semantic links (e.g., Technician–-Robot) regardless of spatial distance, and connect all physical entities to their corresponding auxiliary evidence nodes.

\noindent\textbf{Rank-2 Cells ($X^2$) for Higher-Order Behavior.}
To capture irreducible group dynamics, we build rank-2 cells incident to the underlying rank-1 edges. 
This includes \textit{person cells} that topologically aggregate a single individual's skeleton, and \textit{functional cells} that encapsulate fine-grained multi-actor events (e.g., \{Surgeon, Robot, Saw, Patient\} complexes) guided by clinical priors. 
This explicit modeling creates a shared topological neighborhood, allowing involved entities to update their features simultaneously based on a group state.

\noindent\textbf{Visual Context and Temporal Linking.}
We ground geometric cells visually by projecting 3D nodes onto multi-view 2D planes and fusing sampled features \cite{zhu2021deformable} via attention gating.
To capture the temporal evolution of the surgical scene, we combine consecutive frames into a spatio-temporal complex, establishing bidirectional temporal edges between identical entities across frames.

\subsection{Implementation Details and Multi-Task Learning}
\label{subsec:implementation}

\noindent
\textbf{Entity Initialization.}
To instantiate our entity nodes without requiring laborious manual annotations, we leverage frozen perception modules. 
We use COMPOSE \cite{wang2026compose} to acquire training-free 3D human pose estimations. 
To extract object parameters and human semantics, we align 2D semantic segmentation predictions, generated by a pre-trained model provided with the dataset, with depth estimations derived from DepthAnythingv3 \cite{lin2025depth}.
Using the camera parameters, we back-project the 2D segmentations into a unified 3D space, which yields 3D bounding boxes and allows us to assign definitive semantic class labels (e.g., tools, roles) to our entity nodes.

\noindent
\textbf{Multi-Task Learning.}
Our model is trained end-to-end to infer high-level clinical objectives from the topological representation. 
We attach MLP heads to the pooled CC embeddings to perform simultaneous multi-task learning for \textit{Next Action Anticipation} and \textit{Robot Phase Prediction}.
For sterility breach detection, we apply rule-based spatial heuristics directly over the 3D entities in our topological structure, flagging a violation when a non-sterile entity (e.g., technician) enters within a critical proximity threshold of a sterile entity (e.g., patient). 

\begin{table}[t]
    \centering
    
    \begin{minipage}[t]{0.65\textwidth}
        \centering
        \caption{Performance evaluation on MM-OR \cite{ozsoy2025mm} measured in Macro F1-Score. Best results are \textbf{emphasized}.}
        \label{tab:main_results}
        \resizebox{\linewidth}{!}{%
        \begin{tabular}{@{} l c c c @{}}
            \toprule
            \multirow{2.5}{*}{\textbf{Method}} & \multicolumn{3}{c}{\textbf{F1-Score} ($\uparrow$)} \\
            \cmidrule(l){2-4}
             & \textbf{Sterility Breach} & \textbf{Next Action} & \textbf{Robot Phase} \\
            \midrule
            
            \multicolumn{4}{@{}l}{\textit{LLM-Based}} \\
            \hspace{1em} MM2SG \cite{ozsoy2025mm} & 55.00 & 35.40 & 56.90 \\
            \midrule
            
            \multicolumn{4}{@{}l}{\textit{Latent \& Relational Models}} \\
            \hspace{1em} Vanilla Transformer & \textbf{76.83} & 34.80 & 65.29 \\
            \hspace{1em} SurgLatentGraph \cite{dhamo2020semantic} & \textbf{76.83} & 37.46 & 64.61 \\
            \hspace{1em} \name (Ours) & \textbf{76.83} & \textbf{41.10} & \textbf{73.53} \\
            \bottomrule
        \end{tabular}%
        }
    \end{minipage}\hfill
    \begin{minipage}[t]{0.33\textwidth}
        \centering
        \caption{Scene graph relation prediction F1 when reducing our topological representation to a string-based format \cite{ozsoy2025mm}.}
        \label{tab:reduction_results}
        \resizebox{\linewidth}{!}{%
        \begin{tabular}{@{} l c @{}}
            \toprule
            \textbf{Method} & \textbf{F1-Score} ($\uparrow$) \\
            \midrule
            MM2SG \cite{ozsoy2025mm} & 52.90 \\
            Ours (Decision Tree) & 43.72 \\
            Ours (Learned Head) & \textbf{61.30} \\
            \bottomrule
        \end{tabular}%
        }
    \end{minipage}
    
    \vspace{-4.5mm}
\end{table}

\section{Experiments and Results}
\vspace{-1.0mm}
\label{sec:experiments}

\noindent
\textbf{Dataset and Evaluation Metrics.}
To evaluate our proposed framework, we conduct all experiments on the multimodal MM-OR dataset \cite{ozsoy2025mm} using its standard train/val/test splits. 
We utilize the macro F1-score across all experiments to maintain consistency with established evaluation protocols \cite{ozsoy2025mm}.

\begin{figure}[ht]
    \centering
    \includegraphics[width=\textwidth]{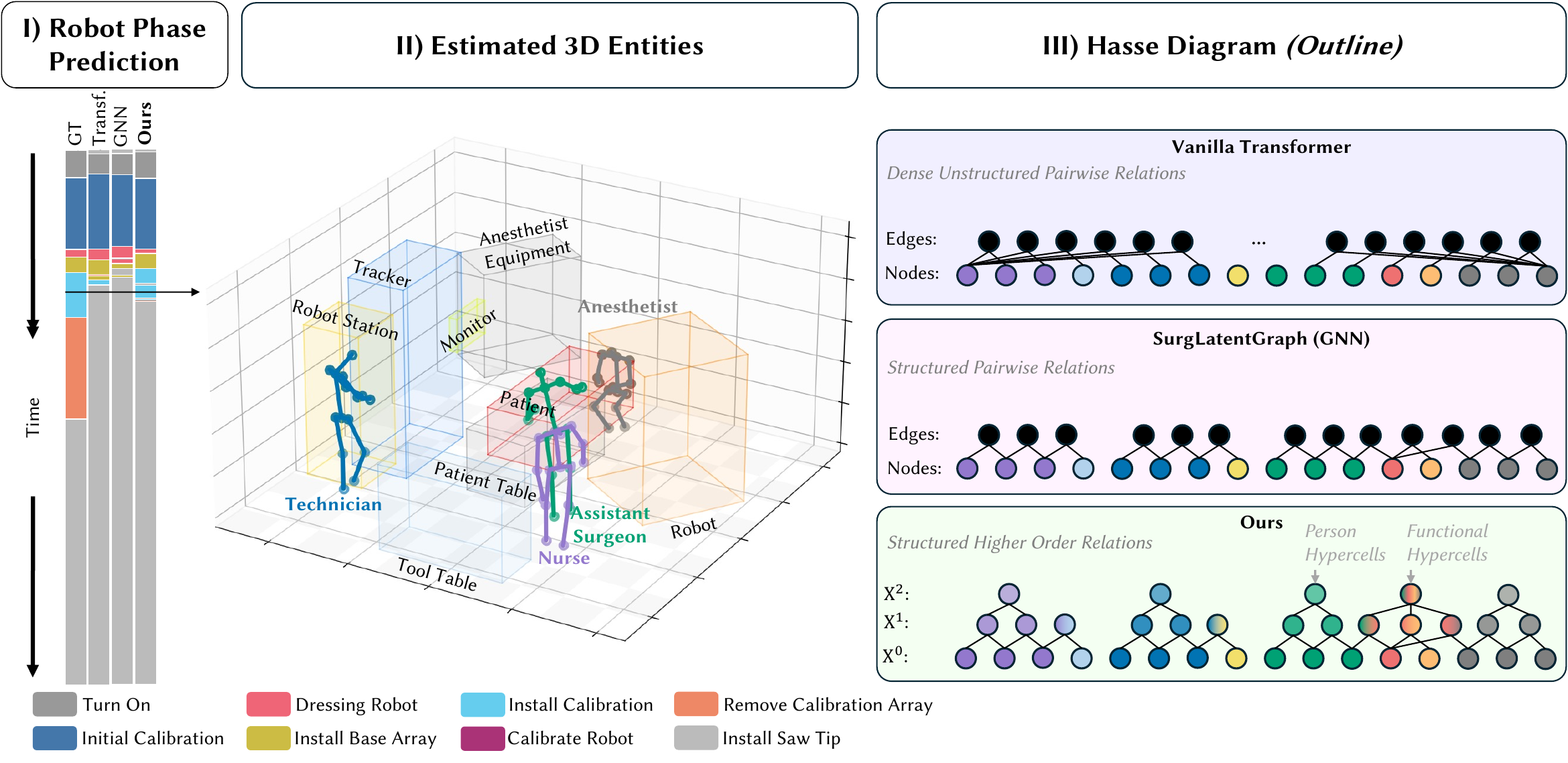}
    \caption{Qualitative results and scene abstraction. 
    We demonstrate improved performance in (I) robot phase prediction over baseline models. 
    While all methods operate on the same (II) explicit 3D input entities, (III) illustrates the distinct structural formulation of each approach.
    Unlike the flattened relations of standard networks, our higher-order representation explicitly models the hierarchical incidence of the surgical environment through rank-0, rank-1, and rank-2 cells.
    \vspace{-3mm}
    }
    \label{fig:qualitative}
\end{figure}

\noindent\textbf{Main Quantitative Results.}
As shown in \cref{tab:main_results}, we compare \name against MM2SG \cite{ozsoy2025mm} (VLM paradigm), a Vanilla Transformer \cite{vaswani2017attention} (structureless sequences), and SurgLatentGraph \cite{dhamo2020semantic} (flattened pairwise graphs).
For sterility breach detection, all 3D-grounded methods achieve 76\% F1, leveraging explicit spatial inputs to significantly outperform the text-reliant MM2SG (55\% F1). 
In next action anticipation, \name achieves 41\%, outperforming the Transformer and SurgLatentGraph by natively preserving irreducible multi-agent dynamics within our rank-2 hypercells. 
Furthermore, \name reaches a state-of-the-art 73\% in robot phase prediction.
Rather than flattening high-frequency 3D poses and textual robot logs into a uniform feature space, \name processes heterogeneous OR cues strictly through topological incidence boundaries, underscoring the advantages of our explicit structural design over flattened architectures.

\noindent
\textbf{Qualitative Results.}
In \cref{fig:qualitative}, we demonstrate our model's superior performance in robot phase prediction.
While shown baselines process identical explicit 3D input entities, they organize and fuse this information differently. 
Unlike Vanilla transformers \cite{vaswani2017attention} (unstructured attention) and standard GNNs \cite{dhamo2020semantic} (pairwise edges), our approach structures the scene as a hierarchical complex, natively capturing the polyadic dynamics required to align with the ground truth.

\noindent
\textbf{Ablation Over Modalities.}
To evaluate the contribution of each modality, we conduct an incremental ablation study, beginning from a purely geometric baseline and systematically introducing additional modalities (\cref{tab:ablation_modalities}).
Relying solely on \textit{estimated} geometric inputs (objects and human skeletons) yields limited performance, with both tasks below 27\% F1.
Grounding these spatial estimates with RGB visual context produces significant improvement, increasing robot phase prediction by over 40\%.
Incrementally fusing heterogeneous non-spatial cues (robot logs, audio) provides consistent additional gains, suggesting that the \name can effectively integrate diverse modalities without degradation.
Finally, temporal edges yield a further improvement, particularly for robot phase prediction, consistent with the expectation that surgical phases are duration-dependent and benefit from temporal context (8 frames).

\begin{table}[t]
    \centering
    \caption{Ablation study of incremental input modalities on Next Action Anticipation and Robot Phase Prediction.}
    \label{tab:ablation_modalities}
    \begin{tabular*}{\textwidth}{@{\extracolsep{\fill}} cccccc cc @{}}
        \toprule
        \multicolumn{6}{c}{\textbf{Input Modalities}} & \multicolumn{2}{c}{\textbf{F1-Score} ($\uparrow$)} \\
        \cmidrule(lr){1-6} \cmidrule(l){7-8}
        
        \textbf{\shortstack{Object \\ Nodes}} & 
        \textbf{\shortstack{Human \\ Skel.}} & 
        \textbf{\shortstack{RGB \\ Image}} & 
        \textbf{\shortstack{Robot \\ Logs}} & 
        \textbf{Audio} & 
        \textbf{Temporality} & 
        \textbf{\shortstack{Next \\ Action}} & 
        \textbf{\shortstack{Robot \\ Phase}} \\
        \midrule
        
        \checkmark & & & & & & 25.82 & 21.08 \\
        \checkmark & \checkmark & & & & & 26.26 & 20.25 \\
        \checkmark & \checkmark & \checkmark & & & & 36.35 & 60.71 \\
        \checkmark & \checkmark & \checkmark & \checkmark & & & 38.26 & 67.06 \\
        \checkmark & \checkmark & \checkmark & \checkmark & \checkmark & & 40.57 & 69.63 \\
        \checkmark & \checkmark & \checkmark & \checkmark & \checkmark & \checkmark & \textbf{41.10} & \textbf{73.53} \\
        \bottomrule
    \end{tabular*}
    \vspace{-4.5mm}
\end{table}

\noindent
\textbf{Representational Power and Graph Reduction.}
To evaluate whether our topological representation subsumes traditional scene graphs, we reduce our continuous combinatorial complex into the discrete string-based format used by \cite{ozsoy2025mm}.
As shown in \cref{tab:reduction_results}, a simple decision tree (depth 16) trained exclusively on rank-0 spatial entities already achieves 43\% F1, indicating that the estimated 3D localization encodes meaningful relational structure.
Training a relation classification head on the full topological complex yields 61\% F1, outperforming the 53\% F1 of the LLM baseline \cite{ozsoy2025mm}, suggesting that the structured multimodal representation retains richer relational information than VLM-generated graphs.

\noindent
\textbf{Runtime Efficiency.}
We compare the parameter count and inference-time efficiency on one NVIDIA A40 GPU. 
\name (12M parameters) requires $59^{\pm 0.45}$ ms per forward pass, while MM2SG \cite{ozsoy2025mm} (7B parameters, 4 bit quantized) requires $194.11^{\pm 2.62}$ ms. The resulting latency advantage makes \name more amenable to intra-operative use, where real-time deployment is essential.

\section{Concluding Remarks}
\vspace{-1.5mm}
\label{sec:conclusion}

We introduced a novel topological framework that redefines the representation of surgical scenes. 
By modeling ORs as multimodal combinatorial complexes, we explicitly preserve the environment's geometric structure and physical reality. 
Unlike standard graph networks or vision-language models that flatten complex interactions, our higher-order attention network (HAT) successfully bridges the gap between atomic entities and collective multi-actor dynamics. 
We show that this structural integrity translates to superior representational expressiveness on the MM-OR dataset. 
Code will be released upon publication.

\noindent
\textbf{Limitations.}
A broader limitation is that current benchmarks for surgical scene understanding focus primarily on downstream classification or regression tasks.
While establishing technical efficacy, these metrics fail to capture the models' real-world clinical utility.
To fully leverage higher-order models, future evaluations should also prioritize clinically actionable metrics like intraoperative risk mitigation, team cognitive load, and context-aware error prevention.

\noindent
\textbf{Acknowledgements}
The authors acknowledge support from the UK AI Research Resource (AIRR) through grant 0251-4584-0945-1 and from the Excellence Strategy of local and state governments in Bavaria, Germany.
T. B. was supported by a UKRI Future Leaders Fellowship (MR/Y018818/1) as well as a Royal Society Research Grant (RG/R1/241402). 
L.B. acknowledges support of the UK Royal Society through grant NIF/R1/254128.

\clearpage
\bibliographystyle{splncs04}
\bibliography{main}

\end{document}